\def\year{2022}\relax
\documentclass[letterpaper]{article} 
\usepackage{aaai22}  
\usepackage{times}  
\usepackage{helvet}  
\usepackage{courier}  
\usepackage[hyphens]{url}  
\usepackage{graphicx} 
\usepackage{natbib}  
\usepackage{caption} 
\DeclareCaptionStyle{ruled}{labelfont=normalfont,labelsep=colon,strut=off} 
\usepackage{algorithm}
\usepackage{algorithmic}

\usepackage{newfloat}
\usepackage{listings}
\floatstyle{ruled}
\newfloat{listing}{tb}{lst}{}
\floatname{listing}{Listing}
\usepackage{tikz}
\usetikzlibrary{fit}
\usepackage{pgfplots}
\usepackage{comment}
\usepackage{subcaption}
\usepackage{booktabs}
\usepackage{multirow}

\title{A*+BFHS: A Hybrid Heuristic Search Algorithm}
\author{
    Zhaoxing Bu, Richard E. Korf
}
\affiliations{

    Computer Science Department\\
    University of California, Los Angeles\\
    Los Angeles, CA 90095\\
    \{zbu, korf\}@cs.ucla.edu
}

\begin{document}

\maketitle

\begin{abstract}
 We present a new algorithm A*+BFHS for solving problems with unit-cost operators where A* and IDA* fail due to memory limitations and/or the existence of many distinct paths between the same pair of nodes. A*+BFHS is based on A* and breadth-first heuristic search (BFHS). 
A*+BFHS combines advantages from both algorithms, namely A*'s node ordering, BFHS's memory savings, and both algorithms' duplicate detection. On easy problems, A*+BFHS behaves the same as A*. 
On hard problems, it is slower than A* but saves a large amount of memory. 
Compared to BFIDA*, A*+BFHS reduces the search time and/or memory requirement by several times on a variety of planning domains.
\end{abstract}

\section{Introduction and Overview}

A* \cite{DBLP:journals/tssc/HartNR68} is a classic heuristic search algorithm that is used by many state-of-the-art optimal track planners \cite{katz2018delfi, DBLP:conf/ijcai/FrancoTLB17, franco2018complementary2, martinez2018planning}. One advantage of A* is duplicate detection. A* uses a Closed list and an Open list to prune duplicate nodes. 
A state is a unique configuration of the problem while a node is a data structure that represents a state reached by a particular path.  
Duplicate nodes represent the same state arrived at via different paths.

The second advantage of A* is node ordering. A* always picks an Open node whose $f$-value is minimum among all Open nodes to expand next, which guarantees an optimal solution returned by A* when using an admissible heuristic. 
When using a consistent heuristic, A* expands all nodes whose $f$-value is less than the optimal solution cost ($C^*$). 
However, tie-breaking among nodes of equal $f$-value significantly affects the set of expanded nodes whose $f$-value equals $C^*$. 
It is common practice to choose an Open node whose $h$-value is minimum among all Open nodes with the same $f$-value, as this strategy usually leads to fewer nodes expanded. A survey of tie-breaking strategies in A* can be found in \cite{DBLP:conf/aaai/AsaiF16}.

A*'s main drawback is its exponential space requirement as it stores in memory all nodes generated during the search. For example, A* can fill up 8 GB of memory in a few minutes on common heuristic search and planning domains. To solve hard problems where A* fails due to memory limitations, researchers have proposed various algorithms, usually by forgoing A*'s duplicate detection or node ordering. For example, Iterative-Deepening-A* (IDA*, \citealt {DBLP:journals/ai/Korf85}) only has a linear memory requirement, at the price of no duplicate detection and a depth-first order within each search bound. However, IDA* may generate too many duplicate nodes on domains containing lots of distinct paths between the same pair of nodes, such as Towers of Hanoi and many planning domains, limiting its application. 
For example, in a grid graph, there are $m+n \choose m$ distinct shortest paths from node ($0$,$0$) to ($m$,$n$), and IDA* cannot detect these as duplicates.

We introduce a new algorithm for solving problems with unit-cost operators, with many distinct paths between the same pair of nodes, where IDA* is not effective. First, we review previous algorithms. Second, we present A*+BFHS, which is based on A* and Breadth-First Heuristic Search \cite{DBLP:conf/aips/ZhouH04}. Third, we present experimental results on 32 hard instances from 18 International Planning Competition (IPC) domains. On those problems, A*+BFHS is slower than A* but requires significantly less memory. Compared to BFIDA*, an algorithm that requires less memory than A*, A*+BFHS reduces the search time and/or memory requirement by several times, and sometimes by an order of magnitude, on a variety of domains.

\section{Previous Work}

IDA* with a transposition table (IDA*+TT, \citealt{DBLP:conf/ijcai/SenB89, DBLP:journals/pami/ReinefeldM94}) uses a transposition table to detect duplicate nodes. However, IDA*+TT is outperformed by other algorithms on both heuristic search \cite{DBLP:conf/ijcai/BuK19} and planning domains \cite{DBLP:conf/aips/ZhouH04}.

A*+IDA* \cite{DBLP:conf/ijcai/BuK19} combines A* and IDA*, and is the state-of-the-art algorithm on the 24-Puzzle. 
It first runs A* until memory is almost full, then runs IDA* below each frontier node without duplicate detection. 
By sorting the frontier nodes with the same $f$-value in increasing order of $h$-values, A*+IDA* can significantly reduce the number of nodes generated in its last iteration. 
Compared to IDA*, 
we 
reported a reduction by a factor of 400 in the total number of nodes generated in the last iteration on all 50 24-Puzzle test cases in \cite{DBLP:journals/ai/KorfF02}. 
Similar to IDA*, A*+IDA* does not work well on domains with many distinct paths between the same pair of nodes.

Frontier search \cite{DBLP:journals/jacm/KorfZTH05} is a family of heuristic search algorithms that work well on domains with many distinct paths between the same pair of nodes. 
Rather than storing all nodes generated, 
it stores only nodes that are at or near the search frontier, including all Open nodes and only one or two layers of Closed nodes. 
As a result, when a goal node is expanded, only the optimal cost is known. 
To reconstruct the solution path, frontier search keeps a middle layer of Closed nodes in memory.
For example, we can save the Closed nodes at depth $h(start)/2$ as the middle layer.  
Each node generated below this middle layer has a pointer to its ancestor in the middle layer. 
After finding the optimal cost, a node in the middle layer that is on an optimal path is identified. 
Then the same algorithm can be applied recursively to compute the solution path from the start node to the middle node, and from the middle node to the goal node. 
In general, however, frontier search cannot prune all duplicates in directed graphs \cite{DBLP:journals/jacm/KorfZTH05,DBLP:conf/aips/ZhouH04}. 

Divide-and-Conquer Frontier-A* (DCFA*, \citealt{DBLP:conf/aaai/KorfZ00}) is a best-first frontier search based on A*. To reconstruct the solution path, DCFA* keeps a middle layer of Closed nodes that are roughly halfway along the solution path. DCFA* detects duplicates and maintains A*'s node ordering, but its memory savings compared to A* is limited on domains where the Open list is larger than the Closed list. 

Breadth-First Heuristic Search (BFHS, \citealt{DBLP:conf/aips/ZhouH04}) is a frontier search algorithm for unit-cost domains. BFHS also detects duplicates but uses a breadth-first node ordering instead of A*'s best-first ordering. At first, assume the optimal cost $C^*$ is known in advance. BFHS runs a breadth-first search (BFS) from the start node and prunes every generated node whose $f$-value exceeds $C^*$. To save memory, BFHS only keeps a few layers of nodes in memory. 
On undirected graphs, if we store the operators used to generate each node, and do not regenerate the parents of a node via the inverses of those operators, frontier search only needs to store two layers of nodes, the currently expanding layer and their child nodes \cite{DBLP:journals/jacm/KorfZTH05}.
On directed graphs, one previous layer besides the above-mentioned two layers is usually stored to detect duplicates \cite{DBLP:conf/aips/ZhouH04}. To reconstruct the solution path, \citeauthor{DBLP:conf/aips/ZhouH04} (\citeyear{DBLP:conf/aips/ZhouH04}) recommend saving the layer at the 3/4 point of the solution length as the middle layer instead of the layer at the halfway point, which usually requires more memory. 
As shown in \cite{DBLP:conf/aips/ZhouH04}, on a domain where the Open list of A* is larger than the Closed list, BFHS usually ends up storing fewer nodes than DCFA*. 

In general, $C^*$ is not known in advance. 
\citeauthor{DBLP:conf/aips/ZhouH04} \citeyearpar{DBLP:conf/aips/ZhouH04} proposed Breadth-First Iterative-Deepening-A* (BFIDA*), which runs
multiple iterations of BFHS, each with a different $f$-bound, starting with the heuristic value of the start node. Similar to IDA*, the last iteration of BFIDA* is often significantly larger than previous iterations, so most search time is often spent on the last iteration.

Compared to A*, BFHS and BFIDA* save significant memory but generate more nodes. 
The main drawback of BFHS and BFIDA* is that their node ordering is almost the worst among different node ordering schemes. BFHS and BFIDA*'s breadth-first ordering means they have to expand all nodes stored at a single depth before expanding any nodes in the next depth. As a result, they have to expand almost all nodes whose $f$-value equals $C^*$, excepting only some nodes at the same depth as the goal node, while A* may only expand a small fraction of such nodes due to its node ordering. 

Forward Perimeter Search (FPS, \citeauthor{DBLP:conf/ijcai/SchuttDR13} \citeyear{DBLP:conf/ijcai/SchuttDR13}) builds a perimeter around the start node via BFS, then runs BFIDA* below each perimeter node. 
The authors only test FPS on the 24-Puzzle and 17-Pancake problem, and did not report any running time.

\section{A*+BFHS}

\subsection{Algorithm Description}

We propose A*+BFHS, a hybrid algorithm to solve problems with many paths between the same pair of nodes. 
A*+BFHS first runs A* until a storage threshold is reached, then runs a series of BFHS iterations on sets of frontier nodes, which are the Open nodes at the end of the A* phase.

The BFHS phase can be viewed as a doubly nested loop. 
Each iteration of the outer loop, which we define as an iteration of the BFHS phase, corresponds to a different cost bound for BFHS. The first cost bound is set to the smallest $f$-value among all frontier nodes. 
In each iteration of the BFHS phase, we first partition the frontier nodes whose $f$-value equals the cost bound into different sets according to their depths. Then the inner loop makes one call to BFHS on each set of frontier nodes, in decreasing order of their depths. This is done by initializing the BFS queue of each call to BFHS with all the nodes in the set. This inner loop continues until a solution is found or all calls to BFHS with the current bound fail to find a solution. 
After each call to BFHS on a set of frontier nodes, we increase the $f$-value of all nodes in the set to the minimum $f$-value of the nodes generated but not expanded in the previous call to BFHS.

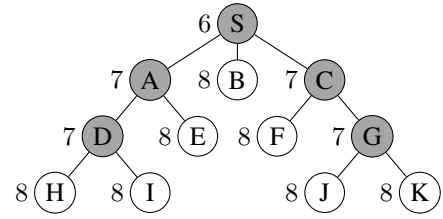
\begin{figure}[t]
\centering
\begin{tikzpicture}[scale=0.5, minimum size=1.5em, inner sep=2pt, sibling distance=6em, label distance=-0.3em]
  \node [fill=gray!70, circle, draw, label=left:$6$] (s) {S} 
    child { node [xshift=-0.3em, fill=gray!70, circle, draw, label=left:$7$] (a1) {A}
      child { node [xshift=-0.3em, fill=gray!70, circle, draw, label=left:$7$] (b1) {D}
        child { node [xshift=-0.3em, circle, draw, label=left:$8$] (c1) {H}
        }
        child { node [xshift=0.3em, circle, draw, label=left:$8$] (c2) {I}
        }
      }
      child { node [xshift=0.3em, circle, draw, label=left:$8$] (b2) {E} }
    }
    child { node [circle, draw, label=left:$8$] (a2) {B}
    }
    child { node [xshift=0.3em, fill=gray!70, circle, draw, label=left:$7$] (a3) {C}
      child { node [xshift=-0.3em, circle, draw, label=left:$8$] (b3) {F} }
      child { node [xshift=0.3em, fill=gray!70, circle, draw, label=left:$7$] (b4) {G}
        child { node [xshift=-0.3em, circle, draw, label=left:$8$] (c3) {J} }
        child { node [xshift=0.3em, circle, draw, label=left:$8$] (c4) {K} }
      }
    }
  ;
  
\end{tikzpicture}
\caption{An example of A*+BFHS's search frontier. Numbers are $f$-values. Closed nodes are gray.}
\label{fig:example_graph}
\end{figure}

Figure \ref{fig:example_graph} presents an example of the Open and Closed nodes at the end of the A* phase. Node S is the start node. All edge costs are 1 and the number next to each node is its $f$-value. Closed nodes are gray. 
The Open nodes B, E, F, H, I, J, K are the frontier nodes for the BFHS phase. 
A*+BFHS first makes a call to BFHS with a cost bound of 8 on all frontier nodes at depth 3, namely nodes H, I, J, K. If no solution is found, A*+BFHS updates the $f$-values of all these nodes to the minimum $f$-value of the nodes generated but not expanded in that call to BFHS. 
A*+BFHS then makes a second call to BFHS with bound 8, starting with all frontier nodes at depth 2, namely nodes E and F. If no solution is found, A*+BFHS updates the $f$-values of these nodes, then makes a third call to BFHS with bound 8, starting with the frontier node B at depth 1. 
Suppose that no solution is found with bound 8, the updated $f$-values for nodes E, F, H, I, J, K are 9, and the updated $f$-value for node B is 10. 
A*+BFHS then starts a new iteration of BFHS with a cost bound of 9, making two calls to BFHS on nodes at depth 3 and 2 respectively. If the solution is found in the first call to BFHS with bound 9, BFHS will not be called again on nodes E and F.
 
A*+BFHS is complete and admissible with admissible heuristics. A*+BFHS potentially makes calls to BFHS on all frontier nodes. When an optimal solution exists, one node on this optimal path serves as one of the start nodes for one of the calls to BFHS. Such a node is guaranteed to exist by A*'s completeness and admissibility. When the cost bound for the calls to BFHS equals $C^*$, the optimal solution will be found, guaranteed by BFHS's completeness and admissibility.

A state can be regenerated in separate calls to BFHS in the same iteration. 
To reduce such duplicates, we can decrease the number of calls to BFHS in each iteration by 
making each call to BFHS on a combined set of frontier nodes at adjacent depths. 
For the example in Figure \ref{fig:example_graph}, we can make one call to BFHS on the frontier nodes at depths 2 and 3 together instead of two separate calls to BFHS, by putting the frontier nodes at depth 3 after the frontier nodes at depth 2 in the initial BFS queue.

In practice, we can specify a maximum number of calls to BFHS per iteration. Then in each iteration, we divide the number of depths of the frontier nodes by the number of calls to BFHS to get the number of depths for each call to BFHS. For example, if the depths of the frontier nodes range from 7 to 12 and we are limited to three calls to BFHS per iteration, each call to BFHS will start with frontier nodes at two depths. We used this strategy in our experiments.

For each node generated in the BFHS phase, we check if it was generated in the A* phase.
If so, we immediately prune the node if its current $g$-value in the BFHS phase is greater than or equal to its stored $g$-value in the A* phase.

The primary purpose of the A* phase is to build a frontier set, so that A*+BFHS can terminate early in its last iteration. In the A* phase we have to reserve some memory for the BFHS phase. In our experiments, we first generated pattern databases or the merge-and-shrink heuristic, then allocated 1/10 of the remaining memory of 8 GB for the A* phase.

\subsection{Comparisons to BFIDA* and FPS} 

A*+BFHS's BFHS phase also uses the iterative deepening concept of BFIDA*, but there are two key differences. First, in each iteration, BFIDA* always makes one call to BFHS on the start node, while we call BFHS multiple times, each on a different set of frontier nodes. Second, in each iteration, 
we order the frontier nodes based on their depth, and run BFHS on the deepest frontier nodes first. 

These differences lead to one drawback and two advantages. The drawback is that A*+BFHS may generate more nodes than BFIDA*, as the same state can be regenerated in separate calls to BFHS in the same iteration. 

The first advantage is that A*+BFHS may terminate early in its last iteration. If A*+BFHS generates a goal node in the last iteration below a relatively deep frontier node, no frontier nodes above that depth will be expanded. Therefore, A*+BFHS may generate only a small number of nodes in its last iteration. In contrast, BFIDA* has to expand almost all nodes whose $f$-value is less than or equal to $C^*$ in its last iteration. As a result, A*+BFHS can be faster than BFIDA*.

The second advantage is that A*+BFHS's memory usage, which is the maximum number of nodes stored during the entire search, may be smaller than that of BFIDA* for two reasons. First, the partition of frontier nodes and separate calls to BFHS within the same iteration can reduce the maximum number of nodes stored in the BFHS phase. Second, BFIDA* stores the most nodes in its last iteration while A*+BFHS may store only a small number of nodes in the last iteration due to early termination. Thus, A*+BFHS may store the most nodes in the penultimate iteration instead. 

FPS looks similar to A*+BFHS, but there are several fundamental differences. First, FPS builds the perimeter using a breadth-first approach while A*+BFHS builds the frontier via a best-first approach. FPS can also dynamically extend the perimeter but this approach does not always speed up the search \cite{DBLP:conf/ijcai/SchuttDR13}. Second, in each iteration of FPS's BFIDA* phase, FPS makes one call to BFHS on each perimeter node. In contrast, in A*+BFHS each call to BFHS is on a set of frontier nodes. Third, FPS sorts the perimeter nodes at the same $f$-value using a max-tree-first or longest-path-first policy, while A*+BFHS sorts the frontier nodes at the same $f$-value in decreasing order of their depth. Fourth, FPS needs two separate searches for solution reconstruction while A*+BFHS only needs one.

\section{Solution Reconstruction}

Each node generated in A*+BFHS's BFHS phase has a pointer to its ancestral frontier node. When a goal node is generated, the solution path from the start node to the ancestral frontier node is stored in the A* phase and only one more search is needed to reconstruct the solution path from the ancestral frontier node to the goal node. This subproblem is much easier than the original problem and we can use the same heuristic function as for the original problem. Therefore, we just use A* to solve this subproblem. In addition, since we know the optimal cost of this subproblem, we can prune any node whose $f$-value exceeds this cost.

In BFIDA*, we have to solve two subproblems to recover the paths from the start node to the middle node and from the middle node to the goal node. \citeauthor{DBLP:conf/aips/ZhouH04} (\citeyear{DBLP:conf/aips/ZhouH04}) called BFHS recursively to solve these two subproblems. 
However, pattern database heuristics (PDB, \citeauthor{DBLP:journals/ci/CulbersonS98} \citeyear{DBLP:journals/ci/CulbersonS98}) only store heuristic values to the goal state, and not between arbitrary pairs of states, which complicates finding a path to a middle node.
Similar to A*+BFHS, we use A* to solve the second subproblem. 
For the first subproblem, we use A* to compute the path from the start node to the middle node using the same heuristic function as for the original problem, which measures the distance to the goal node, not the middle node. To save memory, we prune any node whose $g$-value is greater than or equal to the depth of the middle node, and any node whose $f$-value exceeds the optimal cost of the original problem. Since a deeper middle layer leads to more nodes stored in this approach, we saved the layer at the 1/4 point of the solution length as the middle layer instead of the 3/4 point. In this way, we do not need to build a new heuristic function for the middle node. In our experiments, the search time for solution reconstruction in BFIDA* was usually less than 1\% of the total search time. 

\section{Experimental Results and Analysis}

We implemented BFIDA*, A*+IDA*, and A*+BFHS in the planner Fast Downward 20.06 \cite{DBLP:journals/jair/Helmert06}, using the existing code for node expansion and heuristic functions. 
A*+BFHS's A* phase reused the existing A* code. 
A* stores all nodes in one hash map. We used the same hash map implementation with the following difference. 
In each call to BFHS in both BFIDA* and A*+BFHS, we saved three layers of nodes for duplicate detection and we created one hash map for each layer of nodes. 
We did this because storing all nodes in one hash map in BFHS involves a lot of overhead, and is more complicated. 
\citeauthor{DBLP:conf/ijcai/SchuttDR13} (\citeyear{DBLP:conf/ijcai/SchuttDR13}) did not test FPS on planning domains and we do not know the optimal perimeter radius and sorting strategy for each domain, so we did not implement FPS.

We solved about 550 problem instances from 32 unit-cost domains.
We present the results of A*, BFIDA*, and A*+BFHS on the 32 hardest instances. All remaining instances were easily solved by A*. We tested two A*+BFHS versions. A*+BFHS ($\infty$) starts each call to BFHS on frontier nodes at a single depth. A*+BFHS (4) makes each call to BFHS on frontier nodes at multiple depths with at most four calls to BFHS in each iteration. 
All tests were run on a 3.33 GHz Intel Xeon X5680 CPU with 236 GB of RAM. We used the landmark-cut heuristic (LM-cut, \citeauthor{DBLP:conf/aips/HelmertD09} \citeyear{DBLP:conf/aips/HelmertD09}) for the \textit{satellite} domain, the merge-and-shrink heuristic (M\&S) with the recommended configuration \cite{DBLP:conf/aaai/SieversWH14,DBLP:conf/aips/SieversWH16,DBLP:conf/socs/Sievers18} for the \textit{tpp} and \textit{hiking14} domains, and the iPDB heuristic with the default configuration \cite{DBLP:conf/aaai/HaslumBHBK07,DBLP:conf/socs/SieversOH12} for all other domains.

We present the results in Table \ref{tab:data1}, where the first 26 instances are solved by A* and are sorted by A*'s running time. The last 6 instances in Table \ref{tab:data1} are those where A* terminated 
without finding a solution due to the limitation of the hash map size in Fast Downward 20.06, and are sorted by BFIDA*'s running time. 
We ran each algorithm until it found the optimal cost and returned an optimal path. 
The first column gives the domain name and the instance ID. 
The second through fifth columns give the maximum number of nodes stored by each algorithm. For A*, this is the number of nodes stored at the end of the search. For BFIDA*, this is the largest sum of the number of nodes stored in all three layers of the search, plus the nodes stored in the 1/4 layer for solution reconstruction. For A*+BFHS, this is the largest number of nodes stored in the BFHS phase plus the number of nodes stored in the A* phase. An underline means the specific algorithm needed more than 8 GB of memory to solve the problem. 
The last four columns are the running time in seconds, including the time for solution reconstruction but excluding the time spent on precomputing the heuristic functions, which is the same for all algorithms. 
For each instance, the smallest maximum number of stored nodes and shortest running time are indicated in boldface.
For the last 6 instances where A* terminated without finding a solution, we report A*'s number of nodes and running time when A* terminated, with a $>$ symbol to indicate such numbers.

For A*+IDA*, we set the stored nodes threshold for the A* phase to A*+BFHS's peak stored nodes. A*+IDA* was faster than A*+BFHS by a factor of 2 on \textit{tidybot11} 18, but was slower than A*+BFHS by around 50\% on \textit{mystery} 14, a factor of 2 on \textit{visitall11} 08-half, 4 on \textit{parking14} 16\_9-04, and 8 on \textit{snake18} 17. Furthermore, A*+IDA* was orders-of-magnitude slower than A*+BFHS on domains such as \textit{blocks}, \textit{depot}, \textit{driverlog}, \textit{hiking14}, \textit{logistics00}, \textit{pipesworld-tankage}, \textit{rovers}, \textit{satellite}, \textit{storage}, \textit{termes18}, and \textit{tpp}. Thus we omit the results of A*+IDA* due to space limitations.

\begin{figure}[t]
\centering
\begin{tikzpicture}[scale=0.9]
\begin{axis}[width=3in, 
legend pos= south east, 
        ylabel = {A* time $/$ A*+BFHS time}, 
        xlabel = {A* peak stored \# $/$ A*+BFHS peak stored \#}
    ]
    \addplot[only marks, every mark/.append style={solid, fill=red}, mark=*, mark size=1.5pt] coordinates {
(5.57,0.59)
(6.99,0.66)
(3.68,0.13)
(3.08,0.49)
(4.76,0.55)
(1.51,0.45)
(1.85,0.14)
(5.40,0.13)
(2.48,0.03)
(2.86,0.34)
(4.36,0.41)
(6.89,0.69)
(6.88,0.75)
(9.55,0.64)
(2.76,0.37)
(2.99,0.25)
(2.84,0.20)
(5.43,0.21)
(1.85,0.35)
(3.61,0.44)
(2.66,0.07)
(2.03,0.35)
(1.89,0.25)
(2.67,0.27)
(5.60,0.23)
(6.30,0.55)
    };
    \addlegendentry{\small A*+BFHS (4)}
        \addplot[only marks, every mark/.append style={solid, fill=green}, mark=triangle*,  mark size=1.7pt] coordinates {
(8.61,0.44)
(10.18,0.55)
(4.54,0.08)
(3.35,0.39)
(5.83,0.36)
(1.74,0.57)
(1.98,0.11)
(6.46,0.09)
(2.51,0.02)
(2.96,0.22)
(4.36,0.14)
(6.89,0.69)
(11.47,0.66)
(12.55,0.55)
(2.93,0.35)
(3.31,0.23)
(3.48,0.12)
(5.69,0.22)
(2.15,0.33)
(4.66,0.45)
(2.59,0.02)
(2.80,0.37)
(2.06,0.21)
(4.33,0.32)
(6.06,0.06)
(11.83,0.47)
    };
    \addlegendentry{\small A*+BFHS ($\infty$)}
    \addplot[only marks, every mark/.append style={solid, fill=yellow}, mark=*,  mark size=1pt] coordinates {

  };
\addplot[dashed,mark=] coordinates {
(1,0)
(1,0.75)
}
node[pos=0.8,pin=50:{$x=1$}] {}
;
\end{axis}
\end{tikzpicture}
\caption{A* vs. A*+BFHS in time and memory.} 
\label{fig:ratios_over_Astar}
\end{figure}
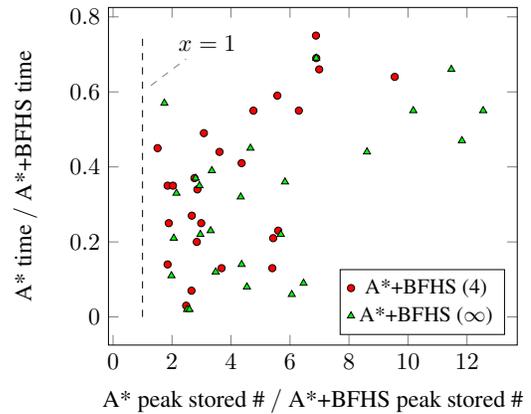

\begin{figure}[t]
\centering
\begin{tikzpicture}[scale=0.9]
\begin{axis}[width=3in, 
legend pos= north east, 
        ylabel = {BFIDA* time $/$ A*+BFHS time}, 
        xlabel = {BFIDA* peak stored \# $/$ A*+BFHS peak stored \#}
    ]
    \addplot[only marks, every mark/.append style={solid, fill=red}, mark=*, mark size=1.5pt] coordinates {
(1.09,0.88)
(1.25,0.92)
(1.07,0.87)
(0.58,0.87)
(0.74,3.59)
(1.16,2.66)
(3.67,2.88)
(1.08,3.36)
(1.33,1.70)
(0.80,1.12)
(0.84,0.55)
(0.77,13.80)
(0.72,11.71)
(6.70,9.09)
(3.59,4.83)
(4.60,3.75)
(5.76,3.84)
(1.94,2.13)
(2.04,1.43)
(1.53,1.54)
(2.37,1.51)
(2.42,1.56)
(1.05,7.09)
(0.86,4.66)
(0.82,2.29)
(2.97,1.68)
(0.31,1.45)
(1.51,1.79)
(1.13,1.37)
(1.74,1.38)
(2.81,2.97)
(2.67,2.22)
    };
    \addlegendentry{\small A*+BFHS (4)}
        \addplot[only marks, every mark/.append style={solid, fill=green}, mark=triangle*,  mark size=1.7pt] coordinates {
(1.68,0.66)
(1.83,0.76)
(1.67,0.63)
(0.72,0.48)
(0.81,2.87)
(1.42,1.78)
(5.13,1.59)
(1.25,4.27)
(1.43,1.34)
(0.96,0.73)
(0.85,0.35)
(0.79,8.95)
(0.72,3.89)
(6.70,9.09)
(5.99,4.25)
(6.04,3.22)
(6.67,3.06)
(2.06,2.00)
(2.25,1.33)
(1.88,0.91)
(2.45,0.60)
(2.43,0.60)
(1.10,7.31)
(1.00,4.32)
(1.06,2.32)
(3.37,1.01)
(0.30,0.31)
(2.08,1.89)
(1.24,1.14)
(2.83,1.63)
(3.04,0.76)
(5.01,1.89)
    }; 
    \addlegendentry{\small A*+BFHS ($\infty$)}
    \addplot[only marks, every mark/.append style={solid, fill=yellow}, mark=*,  mark size=1pt] coordinates {

  };
\addplot[dashed,mark=] coordinates {
(0,1)
(6.7,1)
}
node[pos=0.7,pin=350:{$y=1$}] {}
;
\addplot[dashed,mark=] coordinates {
(1,0)
(1,14)
}
node[pos=0.7,pin=350:{$x=1$}] {}
;
\end{axis}
\end{tikzpicture}
\caption{BFIDA* vs. A*+BFHS in time and memory.}
\label{fig:ratios_over_BFIDAstar}
\end{figure}
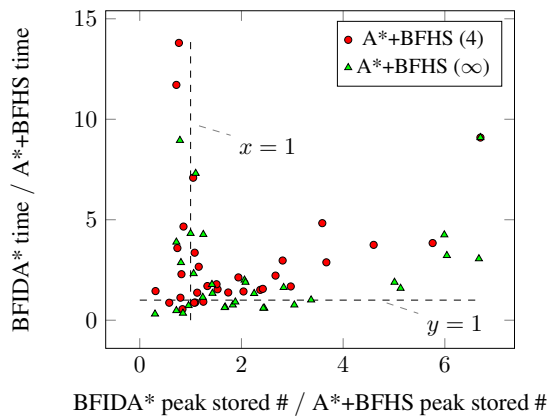

We further compare the time and memory between A* and A*+BFHS in Figure \ref{fig:ratios_over_Astar}, and between BFIDA* and A*+BFHS in Figure \ref{fig:ratios_over_BFIDAstar}, where the $x$-axis is A*/BFIDA*'s peak stored nodes over A*+BFHS's and the $y$-axis is A*/BFIDA*'s running time over A*+BFHS's. Figure \ref{fig:ratios_over_Astar} contains the 26 instances solved by A* and Figure \ref{fig:ratios_over_BFIDAstar} contains all 32 instances. The red circles and green triangles correspond to A*+BFHS (4) and A*+BFHS ($\infty$) respectively. The data points above the $y\!=\!1$ line or to the right of the $x\!=\!1$ line represent instances where A*+BFHS outperformed the comparison algorithm in terms of time or memory.

\begin{table*}[t]
\fontsize{9.0pt}{10.5pt}\selectfont
\centering
\begin{tabular}{c|rrrr|rrrr}  
\toprule
 & \multicolumn{4}{c|}{Peak Stored Nodes} & \multicolumn{4}{c}{Time (s)}\\
 & & & \multicolumn{2}{|c|}{A*+BFHS} & & & \multicolumn{2}{|c}{A*+BFHS}\\
Instance & A* & BFIDA* & \multicolumn{1}{|c}{($\infty$)} & \multicolumn{1}{c|}{(4)} & A* & BFIDA* & \multicolumn{1}{|c}{($\infty$)} & \multicolumn{1}{c}{(4)}\\

\specialrule{0.05em}{0ex}{0.35ex}
\textit{depot} 14 & 70,504,763 & \textbf{17,042,841} & 21,023,657 & 22,882,537 & \textbf{233} & 1,708 & 596 & 475\\

\textit{termes18} 05 & 80,012,545 & \textbf{9,370,587} & 30,874,300 & 30,076,170 & \textbf{245} & 4,796 & 15,415 & 3,319\\

\textit{freecell} 06 & 53,080,996 & 38,054,162 & \textbf{30,481,377} & 35,120,076 & \textbf{250} & 1,883 & 441 & 561\\

\textit{logistics00} 14-1 & 57,689,357 & \textbf{15,441,813} & 19,472,255 & 20,169,648 & \textbf{255} & 10,381 & 1,160 & 752\\

\textit{driverlog} 12 & 144,065,288 & 35,034,406 & \textbf{24,712,720} & 30,270,816 & \textbf{344} & 1,676 & 944 & 631\\

\textit{freecell} 07 & 107,183,015 & 77,196,602 & \textbf{54,171,433} & 58,058,327 & \textbf{522} & 6,416 & 4,775 & 3,769\\

\textit{depot} 11 & \underline{172,447,963} & \textbf{27,192,174} & 37,977,775 & 46,923,423 & \textbf{550} & 3,544 & 7,314 & 4,078\\

\textit{tpp} 11 & \underline{187,011,066} & 93,759,836 & \textbf{30,856,159} & 33,368,912 & \textbf{562} & 7,214 & 9,550 & 2,426\\

\textit{mystery} 14 & \underline{139,924,686} & \underline{135,963,227} & \textbf{20,302,860} & \textbf{20,302,860} & \textbf{578} & 7,628 & 839 & 839\\

\textit{tidybot11} 17 & 69,953,936 & 42,080,838 & \textbf{33,969,968} & 37,090,062 & \textbf{662} & 3,684 & 3,223 & 2,694\\

\textit{logistics00} 15-1 & 82,161,805 & \textbf{13,638,319} & 18,827,830 & 18,827,830 & \textbf{663} & 19,062 & 4,897 & 1,627\\

\textit{pipesworld-notankage} 19 & \underline{123,553,926} & 86,818,434 & \textbf{42,192,503} & 44,706,153 & \textbf{727} & 4,140 & 2,072 & 1,942\\

\textit{parking14} 16\_9-01 & \underline{351,976,816} & 183,832,715 & \textbf{30,675,587} & 51,147,740 & \textbf{971} & 6,236 & 1,468 & 1,290\\

\textit{visitall11} 08-half & \underline{407,182,291} & 172,474,497 & \textbf{34,406,966} & 64,671,078 & \textbf{1,045} & 4,220 & 2,233 & 1,902\\

\textit{tidybot11} 16 & \underline{115,965,857} & 86,095,996 & \textbf{41,342,908} & 57,026,598 & \textbf{1,086} & 5,512 & 2,923 & 3,080\\

\textit{snake18} 08 & 94,699,640 & 44,231,998 & \textbf{44,081,853} & 51,166,308 & \textbf{1,131} & 14,877 & 3,445 & 3,192\\

\textit{hiking14} 2-2-8 & \underline{287,192,625} & \textbf{42,570,885} & 44,454,322 & 53,148,260 & \textbf{1,297} & 10,847 & 14,897 & 9,696\\

\textit{pipesworld-tankage} 14 & \underline{292,998,092} & 158,262,429 & \textbf{84,077,693} & 103,288,306 & \textbf{1,364} & 10,609 & 11,622 & 6,896\\

\textit{blocks} 13-1 & \underline{555,864,249} & 99,782,317 & \textbf{54,601,577} & 79,572,108 & \textbf{1,540} & 2,142 & 2,817 & 2,317\\

\textit{parking14} 16\_9-03 & \underline{606,117,759} & \underline{291,822,896} & \textbf{48,304,204} & 63,455,874 & \textbf{1,714} & 10,059 & 3,124 & 2,679\\

\textit{tidybot11} 18 & \underline{175,574,760} & 114,747,861 & \textbf{40,540,308} & 65,784,369 & \textbf{1,730} & 8,810 & 5,410 & 6,365\\

\textit{blocks} 13-0 & \underline{704,938,102} & 137,821,868 & \textbf{81,918,224} & 126,629,640 & \textbf{1,990} & 2,977 & 4,483 & 3,378\\

\textit{hiking14} 2-3-6 & \underline{368,433,117} & \textbf{124,686,777} & 146,623,619 & 148,357,537 & \textbf{2,480} & 42,379 & 120,494 & 76,603\\

\textit{pipesworld-notankage} 20 & \underline{442,232,520} & \underline{301,349,348} & \textbf{\underline{133,708,317}} & \underline{148,029,967} & \textbf{2,693} & 15,245 & 11,499 & 10,629\\

\textit{snake18} 17 & \underline{265,033,991} & 60,041,363 & \textbf{56,839,243} & 73,365,792 & \textbf{3,967} & 20,418 & 8,785 & 8,916\\

\textit{satellite} 08 & 107,395,076 & 20,846,202 & \textbf{18,870,254} & 19,763,323 & \textbf{11,834} & 398,884 & 54,551 & 56,296\\

\textit{blocks} 15-0 & \underline{$>$814,951,324} & 113,471,990 & \textbf{68,070,197} & 106,482,059 & $>$2,284 & \textbf{3,058} & 4,889 & 3,514\\

\textit{storage} 17 & \underline{$>$799,907,374} & \underline{397,798,456} & \textbf{118,138,352} & 133,800,503 & $>$2,358 & 19,086 & 18,914 & \textbf{11,354}\\

\textit{driverlog} 15 & \underline{$>$786,467,847} & 453,643,579 & \textbf{88,449,751} & 123,602,679 & $>$1,853 & 24,297 & 15,311 & \textbf{8,447}\\

\textit{rovers} 09 & \underline{$>$801,124,989} & 235,386,020 & \textbf{96,100,365} & 99,498,513 & $>$2,776 & 25,336 & 42,290 & \textbf{16,770}\\

\textit{rovers} 11 & \underline{$>$766,016,316} & 274,612,697 & \textbf{112,783,085} & 113,594,902 & $>$2,378 & 26,022 & 43,538 & \textbf{16,661}\\

\textit{parking14} 16\_9-04 & \underline{$>$770,874,998} & \underline{1,045,614,854} & \textbf{156,758,802} & 181,535,647 & $>$2,306 & 37,701 & 12,304 & \textbf{9,813}\\

\addlinespace[-1pt]
\bottomrule
\end{tabular}
\caption{Instances sorted by A* running time if solved by A*. Instances where A* terminated without solving the problem (marked by $>$) are sorted by BFIDA* running time. An underline means more than 8 GB of memory was needed. Smallest memory and shortest times are in boldface.} 
\label{tab:data1}
\end{table*}

\subsection{A*+BFHS vs. A*}

A* was the fastest on all problem instances that it solved, but also used the most memory. 
Among the 32 hardest problem instances we present, A* required more than 8 GB of memory on 22 instances and could not find a solution on 6 of those after running out of the hash map used by Fast Downward 20.06. On some of these instances, A* used 30 GB to 40 GB of memory before it terminated. This means A* cannot solve these 22 instances under the current IPC memory requirement, which is 8 GB.
A*+BFHS required several times, sometimes an order of magnitude, less memory than A*. 
As a result, A*+BFHS only used more than 8 GB of memory on one instance. 
An interesting comparison is the space and time trade-off. For example, on \textit{parking14}, A*+BFHS increased the running time by less than 100\% while saving more than an order of magnitude in memory.

\subsection{A*+BFHS vs. BFIDA*}

In summary, on easy problems that A*+BFHS can solve in its A* phase, A*+BFHS behaves the same as A*, and is always faster than BFIDA*. We solved around 500 such problems, which are not included here due to space limitations. 
On the 32 hardest problems we present, A*+BFHS is faster than BFIDA* on 27 instances and at least twice as fast on 16 of those. Furthermore, A*+BFHS requires less memory than BFIDA* on 25 of the 32 instances and saves more than half the memory on 14 of those. 
In addition, these time and memory reductions exist on both the relatively easy and hard ones of the 32 instances presented, demonstrating that A*+BFHS is in general better than BFIDA* on very hard problems as well as easy problems. 
In the following paragraphs, we compare A*+BFHS with BFIDA* in four aspects: duplicate detection, node ordering, memory, and running time.

Figure \ref{fig:previous_iterations} compares the number of nodes generated prior to the last iteration of BFIDA* and A*+BFHS. For BFIDA*, this is the number of nodes generated in all but the last iteration. For A*+BFHS, this is the sum of nodes generated in its A* phase and all but the last iteration in its BFHS phase. 
The $y$-axis in Figure \ref{fig:previous_iterations} is the number of nodes generated in BFIDA*'s previous iterations divided by A*+BFHS's.
We sort the 32 instances on the $x$-axis according to BFIDA*'s running time, so the left-most instance is the easiest and the right-most instance is the hardest for BFIDA*. The data points above the $y\!=\!1$ line represent instances where A*+BFHS generated fewer nodes than BFIDA* in the previous iterations. 
Compared to BFIDA*, A*+BFHS (4) generated a similar number of nodes in the previous iterations on most instances. \textit{Hiking14} 2-3-6 is the only instance where A*+BFHS (4) generated at least twice as many nodes in the previous iterations as BFIDA*. However, A*+BFHS ($\infty$) generated 2 to 7 times as many nodes in the previous iterations as BFIDA* on 11 instances. This contrast shows that, compared to BFIDA*, significantly more duplicate nodes can be generated by making each call to BFHS on frontier nodes at a single depth. However, most of those duplicate nodes can be avoided by making each call to BFHS on frontier nodes at multiple depths.

A*+BFHS can generate fewer duplicate nodes than BFIDA* due to fewer BFHS iterations and making each call to BFHS on a set of frontier nodes. 
A*+BFHS reduced the number of nodes in previous iterations by around 50\% on \textit{freecell} 06 and \textit{snake18} 17, and a factor of 4 on \textit{snake18} 08. 
To our surprise, we found that on \textit{snake18} 08, the number of nodes generated in the penultimate iteration of BFIDA* was twice as many as the sum of the nodes generated in A*+BFHS's A* phase and the penultimate iteration of the BFHS phase. 
This means a lot of duplicate nodes were generated in BFIDA*. 
\textit{Snake18} generates a directed graph, in which case frontier search cannot detect all duplicate nodes \cite{DBLP:journals/jacm/KorfZTH05,DBLP:conf/aips/ZhouH04}. 

\begin{figure}[t]
\centering
\begin{tikzpicture}[scale=0.9]
\begin{axis}[width=3in, 
legend pos= north west, 
        ylabel = {\small BFIDA*'s previous iterations $/$ A*+BFHS's}, 
        xlabel = {Test instances sorted by BFIDA* running time}, 
    ]
\addplot[only marks, every mark/.append style={solid, fill=red}, mark=*, mark size=1.5pt] coordinates {
(1,1.057423287)
(2,1.347160715)
(3,2.42222806)
(4,1.154505054)
(5,1.110379549)
(6,1.177607647)
(7,0.998644519)
(8,1.225100666)
(9,1.112196515)
(10,1.015616524)
(11,1.503570559)
(12,1.256008973)
(13,1.012530036)
(14,1.527834848)
(15,0.587287276)
(16,1.075805223)
(17,1.173668449)
(18,0.90167418)
(19,0.966165105)
(20,1.142853682)
(21,0.912371569)
(22,4.359231174)
(23,0.874206588)
(24,0.825894381)
(25,0.649625518)
(26,1.885712819)
(27,0.711839552)
(28,0.567718925)
(29,0.557680932)
(30,0.829345281)
(31,0.443530011)
(32,1.084855705)
    };
\addlegendentry{\small A*+BFHS (4)}
\addplot[only marks, every mark/.append style={solid, fill=green}, mark=triangle*,  mark size=1.7pt] coordinates {
(1,0.665069605)
(2,1.07689872)
(3,2.343003932)
(4,1.088291754)
(5,1.001090839)
(6,1.054246139)
(7,0.407637141)
(8,1.041164284)
(9,1.02494361)
(10,0.812905634)
(11,0.318960273)
(12,1.145056645)
(13,0.85932462)
(14,1.222457284)
(15,0.137149092)
(16,1.075805223)
(17,1.062838483)
(18,0.771486485)
(19,0.302011061)
(20,0.638604803)
(21,0.46629868)
(22,3.883158902)
(23,0.729896625)
(24,0.22698348)
(25,0.331505376)
(26,1.681337988)
(27,0.377716164)
(28,0.211466031)
(29,0.204561217)
(30,0.643535541)
(31,0.274125362)
(32,0.942995799)
    };
    \addlegendentry{\small A*+BFHS ($\infty$)}
\addplot[dashed,mark=] coordinates {
(1,1)
(32,1)
}
node[pos=0.7,pin=10:{$y=1$}] {}
;
\end{axis}
\end{tikzpicture}
\caption{The number of nodes generated in BFIDA*'s previous iterations vs. A*+BFHS's.}
\label{fig:previous_iterations}
\end{figure}
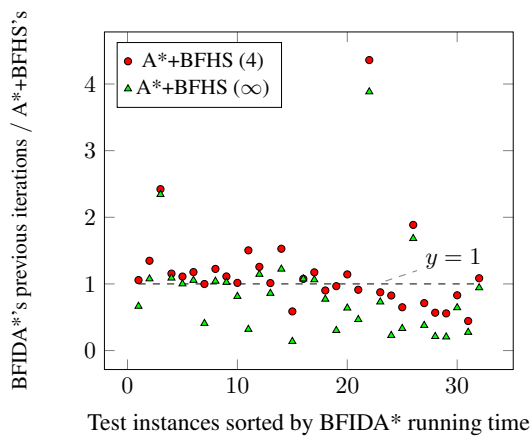

\begin{figure}[t]
\centering
\begin{tikzpicture}[scale=0.9]
\begin{semilogyaxis}[width=3in, 
        ylabel = {BFIDA*'s last iteration $/$ A*+BFHS's}, legend style={at={(0.5,1.15)},anchor=north, /tikz/every even column/.append style={column sep=0.5cm}}, legend columns=-1, xlabel = {Test instances sorted by BFIDA* running time}, 
    ]
\addplot[only marks, every mark/.append style={solid, fill=red}, mark=*, mark size=1.5pt] coordinates {
(1,532.821263)
(2,57.11164056)
(3,4.810589316)
(4,0.869807735)
(5,0.855503905)
(6,0.852091805)
(7,0.853033878)
(8,1.866488747)
(9,127.9912673)
(10,53.11582565)
(11,171.2901597)
(12,5.490929214)
(13,1499.258501)
(14,2.140548418)
(15,26.35956395)
(16,99.63886696)
(17,2.245204175)
(18,39.4469618)
(19,25.76674979)
(20,20.8279464)
(21,1.576927067)
(22,36.24152246)
(23,7.813922229)
(24,49.80474391)
(25,8.207058127)
(26,8.152519137)
(27,98.55676869)
(28,144.4540121)
(29,22.37656453)
(30,15421.01406)
(31,3.018645465)
(32,47.06923839)
    };
\addlegendentry{\small A*+BFHS (4)}
\addplot[only marks, every mark/.append style={solid, fill=green}, mark=triangle*,  mark size=1.7pt] coordinates {
(1,36846.09872)
(2,49.99007172)
(3,33.72595265)
(4,0.670035665)
(5,0.600820867)
(6,0.555105741)
(7,0.552709261)
(8,1.448356026)
(9,7106.039767)
(10,32704.86593)
(11,242.5857581)
(12,32.64921469)
(13,69134.81969)
(14,1.577021417)
(15,76448.91447)
(16,99.63886696)
(17,37.14128726)
(18,32.87693161)
(19,8194.130661)
(20,332.1242841)
(21,1.436312875)
(22,989.3484907)
(23,34.96096268)
(24,342.726138)
(25,224.10566)
(26,96.79629969)
(27,1237.374939)
(28,108432.0351)
(29,3490.955043)
(30,13770.2342)
(31,11.90129523)
(32,13556.53824)
    };
    \addlegendentry{\small A*+BFHS ($\infty$)}
\addplot[dashed,mark=] coordinates {
(1,1)
(32,1)
}
node[pos=0.7,pin=350:{$y=1$}] {}
;
\end{semilogyaxis}
\end{tikzpicture}
\caption{The number of nodes generated in BFIDA*'s last iteration vs. A*+BFHS's.}
\label{fig:last_iteration} 
\end{figure}
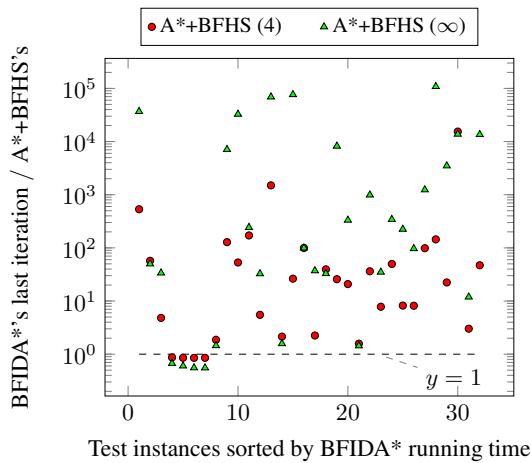

Compared to BFIDA*, A*+BFHS reduced the number of nodes in the last iteration significantly, and usually by several orders of magnitude, on 28 of the 32 instances. 
We present this comparison in Figure \ref{fig:last_iteration}, where the $y$-axis is the number of nodes generated in BFIDA*'s last iteration divided by A*+BFHS's. We also sort the 32 instances on the $x$-axis according to BFIDA*'s running time, so the left-most instance is the easiest and the right-most instance is the hardest for BFIDA*. The data points above the $y\!=\!1$ line represent instances where A*+BFHS generated fewer nodes than BFIDA* in the last iteration. 
Both A*+BFHS versions usually generated several orders of magnitude fewer nodes in the last iteration than BFIDA*, while A*+BFHS ($\infty$) generated the fewest nodes on most instances.
This large reduction proves that when ordering the frontier nodes by deepest-first, A*+BFHS can terminate early in its last iteration. 
On the three \textit{blocks} instances and \textit{depot} 11, A*+BFHS did not terminate early in its last iteration because the ancestral frontier node of the goal had a relatively low $g$-value. 
In fact, A* generated the most nodes while expanding the Open nodes whose $f \!=\! C^*$ on the three \textit{blocks} instances, which shows that node ordering is also difficult for A* on those instances.
In contrast, A* generated very few nodes while expanding the Open nodes whose $f \!=\! C^*$ on \textit{depot} 11, suggesting that A*+BFHS may terminate early in its last iteration given more memory for its A* phase.

A*+BFHS's A* phase usually stored from 10 to 20 million nodes, with the exception of the \textit{snake18} domain where 40 to 50 million nodes were stored. 
Comparing the maximum number of stored nodes, A*+BFHS ($\infty$) required less memory than BFIDA* on 25 instances and less than half the memory on 14 of those. 
For A*+BFHS (4), these two numbers are 23 and 11 respectively.
In contrast, \textit{termes18} 05 is the only instance where the maximum number of stored nodes of A*+BFHS was at least twice that of BFIDA*.

Comparing the two versions of A*+BFHS, A*+BFHS (4) was usually faster, sometimes significantly, due to the reduction in duplicate nodes. Compared to BFIDA*, A*+BFHS (4) was slightly slower on four instances and 80\% slower on one instance. 
On the other 27 instances, A*+BFHS (4) was faster than BFIDA*, and at least twice as fast on 16 of those. 
The large speedups usually were on the instances where BFIDA* generated the most nodes in its last iteration. The best result was on the \textit{logistics00} domain, where an order of magnitude speedup was achieved. This is because BFIDA* performed very poorly on this domain due to its breadth-first node ordering. 
Compared to BFIDA*, 
A*+BFHS ($\infty$) was slower on 11 instances and at least twice as slow on three of those, but also at least twice as fast on 12 instances. The main reason for the slower cases is the presence of many duplicate nodes generated in certain domains. 

\subsection{Calling BFHS on Nodes at Multiple Depths} 
Comparing the two A*+BFHS versions, each has its pros and cons. 
A*+BFHS (4) always generated fewer duplicate nodes. Comparing the number of nodes generated in the previous iterations, A*+BFHS ($\infty$) generated at least twice as many nodes on 7 instances. 
A*+BFHS ($\infty$) generated significantly fewer nodes in the last iteration than A*+BFHS (4) on 22 instances. 
However, the number of nodes generated in the last iteration of A*+BFHS is usually only a small portion of the total nodes generated, so the large difference in the last iteration is not very important. 
A*+BFHS (4) stored a larger maximum number of nodes than A*+BFHS ($\infty$) on almost all instances. However, the difference was usually small and never more than a factor of two. 
For the running time, the difference was usually less than 50\%. Compared to A*+BFHS ($\infty$), A*+BFHS (4) was faster by a factor of 3 on \textit{logistics00} 15-1, 2.5 on \textit{rovers} 09 and 11, 4.6 on \textit{termes18} 05, 3.9 on \textit{tpp} 11, and never more than 30\% slower. 

In general, making each call to BFHS on frontier nodes at multiple depths increases both the memory usage and the number of nodes generated in the last iteration, but reduces the number of duplicate nodes and hence is often faster. Considering the memory and time trade-off, given a new problem, we recommend making each call to BFHS on frontier nodes at multiple depths. However, if we limit the number of calls to BFHS in each iteration to one, then A*+BFHS (1) will generate about the same number of nodes as BFIDA*, and early termination is no longer possible. Therefore, at least two calls should be used. So far, we have only limited BFHS to four calls in each iteration. Determining the optimal number of calls to BFHS is a subject for future work. 

\subsection{Heuristic Functions and Running Time}

For each node generated, A* first does duplicate checking then looks up its heuristic value if needed. Thus for each state, A* only computes its heuristic value once, no matter how many times this state is generated. However, the situation is different in BFHS. Even in a single call to BFHS, a state's heuristic value may be calculated multiple times. For example, if a state's $f$-value is greater than the cost bound of BFHS, then this state is never stored in this call to BFHS and its heuristic value has to be computed every time it is generated. In addition, A* has only one hash map but our BFHS implementation has one hash map for each layer of nodes. Consequently, for each node generated, A* does only one hash map lookup while BFHS may have multiple lookups. 

Due to the above differences, the number of nodes generated per second of BFIDA* and A*+BFHS was smaller than that of A*. For the iPDB and M\&S heuristics, this difference was usually less than a factor of two. For the LM-cut heuristic, A* was faster by a factor of four in terms of nodes generated per second on the \textit{satellite} domain. This is because computing a node's LM-cut heuristic is much more expensive than iPDB and M\&S heuristics. This contrast shows that the choice of heuristic function also plays an important role in comparing the running time of different algorithms. 

\section{Future Work}

Future work includes the following. 
First, test A*+BFHS on more unit-cost domains. 
Second, investigate what is the best memory threshold for the A* phase. 
Third, determine the optimal number of calls to BFHS in each iteration. 
Fourth, find other ways to partition the frontier nodes besides the current depth-based approach. 
If a set of frontier nodes is too large, we may split it into multiple smaller sets and make one call to BFHS on each such smaller set. This approach may reduce the maximum number of stored nodes but may generate more duplicate nodes. In addition, when we make each call to BFHS on frontier nodes at multiple depths, we may consider the number of frontier nodes at each depth so each call to BFHS is on a different number of depths instead of a fixed number. 
Fifth, use external memory such as magnetic disk or flash memory in A*+BFHS to solve very hard problems. For example, instead of allocating 1/10 of RAM for the A* phase, we can first run A* until RAM is almost full, then store both Open and Closed nodes in external memory and remove them from RAM. Then in the BFHS phase, we load back the set of frontier nodes for each call to BFHS from external memory. 
This A*+BFHS version would never perform worse than A*, since it is identical to A* until memory is exhausted, at which point the BHFS phase would begin. 

\section{Conclusions}

We introduce a hybrid heuristic search algorithm A*+BFHS for solving  problems with unit-cost operators that cannot be solved by A* due to memory limitations, nor IDA* due to the existence of many distinct paths between the same pair of nodes. A*+BFHS first runs A* until a user-specified storage threshold is reached, then runs multiple iterations of BFHS on the frontier nodes, which are the Open nodes at the end of the A* phase. Each iteration has a unique cost bound and contains multiple calls to BFHS. Each call to BFHS within the same iteration has the same cost bound but a different set of frontier nodes to start with. Within an iteration, frontier nodes are sorted deepest-first so that A*+BFHS can terminate early in its last iteration.

On the around 500 easy problems solved, A*+BFHS behaves the same as A*, and is always faster than BFIDA*. 
On the 32 hard instances presented, A*+BFHS is slower than A* but uses significantly less memory. 
A*+BFHS is faster than BFIDA* on 27 of those 32 instances and at least twice as fast on 16 of those. 
Furthermore, A*+BFHS requires less memory than BFIDA* on 25 of those 32 instances and saves more than half the memory on 14 of those. Another contribution of this paper is a comprehensive testing of BFIDA* on many planning domains, which is lacking in the literature.

\bibliography{zburef}

\begin{thebibliography}{23}
\providecommand{\natexlab}[1]{#1}

\bibitem[{Asai and Fukunaga(2016)}]{DBLP:conf/aaai/AsaiF16}
Asai, M.; and Fukunaga, A.~S. 2016.
\newblock Tiebreaking Strategies for A* Search: How to Explore the Final
  Frontier.
\newblock In \emph{Proceedings of the Thirtieth {AAAI} Conference on Artificial
  Intelligence, February 12-17, 2016, Phoenix, Arizona, {USA}}, 673--679.
  {AAAI} Press.

\bibitem[{Bu and Korf(2019)}]{DBLP:conf/ijcai/BuK19}
Bu, Z.; and Korf, R.~E. 2019.
\newblock A*+IDA*: {A} Simple Hybrid Search Algorithm.
\newblock In \emph{Proceedings of the Twenty-Eighth International Joint
  Conference on Artificial Intelligence, {IJCAI} 2019, Macao, China, August
  10-16, 2019}, 1206--1212. ijcai.org.

\bibitem[{Culberson and Schaeffer(1998)}]{DBLP:journals/ci/CulbersonS98}
Culberson, J.~C.; and Schaeffer, J. 1998.
\newblock Pattern Databases.
\newblock \emph{Comput. Intell.}, 14(3): 318--334.

\bibitem[{Franco et~al.(2018)Franco, Lelis, Barley, Edelkamp, Martines, and
  Moraru}]{franco2018complementary2}
Franco, S.; Lelis, L.~H.; Barley, M.; Edelkamp, S.; Martines, M.; and Moraru,
  I. 2018.
\newblock The Complementary2 planner in the IPC 2018.
\newblock \emph{IPC-9 planner abstracts}, 28--31.

\bibitem[{Franco et~al.(2017)Franco, Torralba, Lelis, and
  Barley}]{DBLP:conf/ijcai/FrancoTLB17}
Franco, S.; Torralba, {\'{A}}.; Lelis, L. H.~S.; and Barley, M. 2017.
\newblock On Creating Complementary Pattern Databases.
\newblock In \emph{Proceedings of the Twenty-Sixth International Joint
  Conference on Artificial Intelligence, {IJCAI} 2017, Melbourne, Australia,
  August 19-25, 2017}, 4302--4309. ijcai.org.

\bibitem[{Hart, Nilsson, and Raphael(1968)}]{DBLP:journals/tssc/HartNR68}
Hart, P.~E.; Nilsson, N.~J.; and Raphael, B. 1968.
\newblock A Formal Basis for the Heuristic Determination of Minimum Cost Paths.
\newblock \emph{{IEEE} Trans. Syst. Sci. Cybern.}, 4(2): 100--107.

\bibitem[{Haslum et~al.(2007)Haslum, Botea, Helmert, Bonet, and
  Koenig}]{DBLP:conf/aaai/HaslumBHBK07}
Haslum, P.; Botea, A.; Helmert, M.; Bonet, B.; and Koenig, S. 2007.
\newblock Domain-Independent Construction of Pattern Database Heuristics for
  Cost-Optimal Planning.
\newblock In \emph{Proceedings of the Twenty-Second {AAAI} Conference on
  Artificial Intelligence, July 22-26, 2007, Vancouver, British Columbia,
  Canada}, 1007--1012. {AAAI} Press.

\bibitem[{Helmert(2006)}]{DBLP:journals/jair/Helmert06}
Helmert, M. 2006.
\newblock The Fast Downward Planning System.
\newblock \emph{J. Artif. Intell. Res.}, 26: 191--246.

\bibitem[{Helmert and Domshlak(2009)}]{DBLP:conf/aips/HelmertD09}
Helmert, M.; and Domshlak, C. 2009.
\newblock Landmarks, Critical Paths and Abstractions: What's the Difference
  Anyway?
\newblock In \emph{Proceedings of the 19th International Conference on
  Automated Planning and Scheduling, {ICAPS} 2009, Thessaloniki, Greece,
  September 19-23, 2009}. {AAAI}.

\bibitem[{Katz et~al.(2018)Katz, Sohrabi, Samulowitz, and
  Sievers}]{katz2018delfi}
Katz, M.; Sohrabi, S.; Samulowitz, H.; and Sievers, S. 2018.
\newblock Delfi: Online planner selection for cost-optimal planning.
\newblock \emph{IPC-9 planner abstracts}, 57--64.

\bibitem[{Korf(1985)}]{DBLP:journals/ai/Korf85}
Korf, R.~E. 1985.
\newblock Depth-First Iterative-Deepening: An Optimal Admissible Tree Search.
\newblock \emph{Artif. Intell.}, 27(1): 97--109.

\bibitem[{Korf and Felner(2002)}]{DBLP:journals/ai/KorfF02}
Korf, R.~E.; and Felner, A. 2002.
\newblock Disjoint pattern database heuristics.
\newblock \emph{Artif. Intell.}, 134(1-2): 9--22.

\bibitem[{Korf and Zhang(2000)}]{DBLP:conf/aaai/KorfZ00}
Korf, R.~E.; and Zhang, W. 2000.
\newblock Divide-and-Conquer Frontier Search Applied to Optimal Sequence
  Alignment.
\newblock In \emph{Proceedings of the Seventeenth National Conference on
  Artificial Intelligence and Twelfth Conference on on Innovative Applications
  of Artificial Intelligence, July 30 - August 3, 2000, Austin, Texas, {USA}},
  910--916. {AAAI} Press / The {MIT} Press.

\bibitem[{Korf et~al.(2005)Korf, Zhang, Thayer, and
  Hohwald}]{DBLP:journals/jacm/KorfZTH05}
Korf, R.~E.; Zhang, W.; Thayer, I.; and Hohwald, H. 2005.
\newblock Frontier search.
\newblock \emph{J. {ACM}}, 52(5): 715--748.

\bibitem[{Martinez et~al.(2018)Martinez, Moraru, Edelkamp, and
  Franco}]{martinez2018planning}
Martinez, M.; Moraru, I.; Edelkamp, S.; and Franco, S. 2018.
\newblock Planning-PDBs planner in the IPC 2018.
\newblock \emph{IPC-9 planner abstracts}, 63--66.

\bibitem[{Reinefeld and Marsland(1994)}]{DBLP:journals/pami/ReinefeldM94}
Reinefeld, A.; and Marsland, T.~A. 1994.
\newblock Enhanced Iterative-Deepening Search.
\newblock \emph{{IEEE} Trans. Pattern Anal. Mach. Intell.}, 16(7): 701--710.

\bibitem[{Sch{\"{u}}tt, D{\"{o}}bbelin, and
  Reinefeld(2013)}]{DBLP:conf/ijcai/SchuttDR13}
Sch{\"{u}}tt, T.; D{\"{o}}bbelin, R.; and Reinefeld, A. 2013.
\newblock Forward Perimeter Search with Controlled Use of Memory.
\newblock In \emph{{IJCAI} 2013, Proceedings of the 23rd International Joint
  Conference on Artificial Intelligence, Beijing, China, August 3-9, 2013},
  659--665. {IJCAI/AAAI}.

\bibitem[{Sen and Bagchi(1989)}]{DBLP:conf/ijcai/SenB89}
Sen, A.~K.; and Bagchi, A. 1989.
\newblock Fast Recursive Formulations for Best-First Search That Allow
  Controlled Use of Memory.
\newblock In \emph{Proceedings of the 11th International Joint Conference on
  Artificial Intelligence. Detroit, MI, USA, August 1989}, 297--302. Morgan
  Kaufmann.

\bibitem[{Sievers(2018)}]{DBLP:conf/socs/Sievers18}
Sievers, S. 2018.
\newblock Merge-and-Shrink Heuristics for Classical Planning: Efficient
  Implementation and Partial Abstractions.
\newblock In \emph{Proceedings of the Eleventh International Symposium on
  Combinatorial Search, {SOCS} 2018, Stockholm, Sweden - 14-15 July 2018}, 99.
  {AAAI} Press.

\bibitem[{Sievers, Ortlieb, and Helmert(2012)}]{DBLP:conf/socs/SieversOH12}
Sievers, S.; Ortlieb, M.; and Helmert, M. 2012.
\newblock Efficient Implementation of Pattern Database Heuristics for Classical
  Planning.
\newblock In \emph{Proceedings of the Fifth Annual Symposium on Combinatorial
  Search, {SOCS} 2012, Niagara Falls, Ontario, Canada, July 19-21, 2012}.
  {AAAI} Press.

\bibitem[{Sievers, Wehrle, and Helmert(2014)}]{DBLP:conf/aaai/SieversWH14}
Sievers, S.; Wehrle, M.; and Helmert, M. 2014.
\newblock Generalized Label Reduction for Merge-and-Shrink Heuristics.
\newblock In \emph{Proceedings of the Twenty-Eighth {AAAI} Conference on
  Artificial Intelligence, July 27 -31, 2014, Qu{\'{e}}bec City, Qu{\'{e}}bec,
  Canada}, 2358--2366. {AAAI} Press.

\bibitem[{Sievers, Wehrle, and Helmert(2016)}]{DBLP:conf/aips/SieversWH16}
Sievers, S.; Wehrle, M.; and Helmert, M. 2016.
\newblock An Analysis of Merge Strategies for Merge-and-Shrink Heuristics.
\newblock In \emph{Proceedings of the Twenty-Sixth International Conference on
  Automated Planning and Scheduling, {ICAPS} 2016, London, UK, June 12-17,
  2016}, 294--298. {AAAI} Press.

\bibitem[{Zhou and Hansen(2004)}]{DBLP:conf/aips/ZhouH04}
Zhou, R.; and Hansen, E.~A. 2004.
\newblock Breadth-First Heuristic Search.
\newblock In \emph{Proceedings of the Fourteenth International Conference on
  Automated Planning and Scheduling {(ICAPS} 2004), June 3-7 2004, Whistler,
  British Columbia, Canada}, 92--100. {AAAI}.

\end{thebibliography}


\end{document}